\documentclass{article}

\usepackage[preprint,nonatbib]{neurips_2024}

\usepackage[utf8]{inputenc} 
\usepackage[T1]{fontenc}    
\usepackage{hyperref}       
\usepackage{url}            
\usepackage{booktabs}       
\usepackage{amsfonts}       
\usepackage{nicefrac}       
\usepackage{microtype}      
\usepackage{xcolor}         
\usepackage{amsmath}
\usepackage{amssymb}
\usepackage{biblatex}
\usepackage{algpseudocode}
\usepackage{algorithm}

\title{Parameter-Efficient Neural CDEs via Implicit Function Jacobians}

\author{%
  Ilya Kuleshov \\
  Applied AI Institute\\
  \And
  Alexey Zaytsev\\
  Applied AI Institute
}

\begin{document}

\maketitle

\begin{abstract}
  Neural Controlled Differential Equations (Neural CDEs, NCDEs) are a unique branch of methods, specifically tailored for analysing temporal sequences.
  However, they come with drawbacks, the main one being the number of parameters, required for the method's operation.
  In this paper, we propose an alternative, parameter-efficient look at Neural CDEs.
  It requires much fewer parameters, while also presenting a very logical analogy as the "Continuous RNN", which the Neural CDEs aspire to.
\end{abstract}

\section{Introduction}
Consider a time series $S=\{(t_i, \mathbf{x}_i)\}_{i=1}^T$ as input, where $t_i \in \mathbb{R}$ are the timestamps and $\mathbf{x}_i \in \mathbb{R}^u$ are the corresponding observations.
Neural Controlled Differential Equations (Neural CDEs, NCDEs) were first introduced in the paper~\cite{kidger2020neural}, as a way to aggregate $S$ into an embedding $\mathbf{h}(T) \in \mathbb{R}^v$, by solving the following Cauchy problem:
\begin{equation} \label{eq:ncde_diffeq}
    \dot{\mathbf{h}}_t = f_\theta(\mathbf{X}_t, \mathbf{h}_t) \dot{\mathbf{X}}_t.
\end{equation}
Here, we introduced the following notation:
\begin{itemize}
    \item The subscript $A_t = A(t)$ denotes the timestamp, and the dot above the symbol $\dot{A} = \frac{d A}{dt}$ denotes the timewise derivative;
    \item $\mathbf{X}_t: \mathbb{R} \rightarrow \mathbb{R}^u$ is some interpolation of the input sequence $S$. For instance, the original paper uses cubic interpolation;
    \item $\mathbf{h}_t: \mathbb{R} \rightarrow \mathbb{R}^v$ is the solution of this differential equation, which also coincides with the output of this layer. We will also refer to this vector-function as the \textit{hidden trajectory}.
    \item $f_\theta: \mathbb{R}^u \times \mathbb{R}^v \rightarrow \mathbb{R}^{u \times v}$ is the Neural Network, which defines the NCDE layer.
    It outputs a \textit{control matrix}, which is then multiplied by the derivative of the input interpolation $\dot{\mathbf{X}}_t$ to get the hidden trajectory derivative $\dot{\mathbf{h}}_t$.
\end{itemize}

Even after this short introduction to Neural CDEs, we may already notice one of its drawbacks.
Since $\dot{\mathbf{h}}$ needs to be a vector, the product $f_\theta \dot{\mathbf{X}}_t$ also needs to be a vector.
Consequently, $f_\theta$ must be a matrix, meaning that the dimensionality of the final set of weights in the Neural Network is $\sim d \times u \times v$, where $d$ is its hidden dimension.
As a consequence, the number of parameters of a Neural CDE is dominated by this final weight matrix, cubic w.r.t. the channel dimensions.

\section{Method}
\subsection{The Truly-Continuous RNN}
One of the ideas behind Neural ODEs is its interpretation as the continuous limits of traditional Neural Networks~\cite{chen2018neural}.
Further works tried to expand this concept towards Recurrent Neural Networks: via Neural CDEs~\cite{kidger2020neural}, or by constructing a difference equation~\cite{de2019gru}.
However, these approaches cannot be considered the true "continuous RNNs``.
Neural CDEs do not actually use an RNN as its backbone, while GRU-ODE constructs a difference equation instead of a differential one.
None of them actually attempt to directly take the RNN difference step to the continuous limit, at $\Delta t \rightarrow 0$.

We will attempt to do just that.
Consider the discrete RNN step (irrespective of the notation we introduced above):
$$
\mathbf{h}_{i} = \text{RNN}(\mathbf{x}_i, \mathbf{h}_{i-1}).
$$
In the continuous-limit analogy, this represents the difference equation for $\Delta t = 1$.
Let's substitute the $\Delta t$, replace the index $i$ with the timestamp $t$, and the discrete sequence $\mathbf{x}_i$ with its interpolation $\mathbf{X}_t$:
$$
\mathbf{h}_{t} = \text{RNN}(\mathbf{X}_t, \mathbf{h}_{t-\Delta t}).
$$
Now all that's left is to take the limit $\Delta t \rightarrow 0$, replacing the discrete $\mathbf{x}_t$:
\begin{equation} \label{eq:implicit-function}
\mathbf{h}_t = \text{RNN}(\mathbf{X}_t, \mathbf{h}_t).
\end{equation}
We are left with the equation~\eqref{eq:implicit-function}, which defines the trajectory~$\mathbf{h}_t$ implicitly.
Below, we will show how one can model such a function efficiently.

\subsection{Implicit Function Calculation}
The idea is to differentiate~\eqref{eq:implicit-function} w.r.t. $t$; we also denote the RNN as $\mathbf{f}$ for brevity:
$$
\dot{\mathbf{h}}_t = \frac{d\mathbf{f}(\mathbf{X}_t, \mathbf{h}_t)}{dt} = \frac{\partial \mathbf{f}}{\partial \mathbf{X}} \dot{\mathbf{X}}_t + \frac{\partial \mathbf{f}}{\partial \mathbf{h}} \dot{\mathbf{h}}_t.
$$
Here we encounter the Jacobians: $J_x \triangleq \frac{\partial \mathbf{f}}{\partial \mathbf{X}}; J_h \triangleq \frac{\partial \mathbf{f}}{\partial \mathbf{h}}$.
We also have a $\dot{\mathbf{h}}_t$ on the right side, which prevents us from solving this equation numerically.
Gathering the terms, we get:
$$
    (I - J_h) \dot{\mathbf{h}}_t = J_x \dot{\mathbf{X}}_t
    \implies
    \dot{\mathbf{h}}_t = (I - J_h) ^{-1} J_x \dot{\mathbf{X}}_t,
$$
where $I$ is the identity matrix.
Unfortunately, inverting $I - J_h$ is a very expensive operation, meaning that we cannot expect to calculate the right-hand side quickly enough to use a numerical solver.

\subsection{Speeding up}
The main obstacle in terms of speed is $(I - J_h) ^{-1}$.
Fortunately, for that term, we can use the Taylor expansion:
$$
(I - J_h) ^{-1} = I + J_h + J_h^2 + \ldots.
$$
Substituting this expansion, we are left with:
\begin{equation} \label{eq:jacde_precise}
    \dot{\mathbf{h}}_t = J_x \dot{\mathbf{X}}_t + J_h J_x \dot{\mathbf{X}}_t + o\left(\| J_h \|^2 \right).
\end{equation}

\begin{algorithm}
\caption{The implicit function Jacobian vector field.}
\begin{algorithmic}
    \State \texttt{INPUT} $t, \mathbf{h}_t$
    \State $\mathbf{X}_t, \dot{\mathbf{X}}_t \gets \text{Interpolation}(t|S)$ \Comment{Calculate both the value and the derivative of the interpolant.}
    \State $\mathbf{g}_x \gets J_x \dot{\mathbf{X}}_t$ \Comment{Calculate the $J_x \dot{\mathbf{X}}_t$ Jacobian-vector product.}
    \State $\mathbf{g}_{xh} \gets J_h \mathbf{g}_x$
    \Comment{Calculate the $J_h J_x \dot{\mathbf{X}}_t$ Jacobian-vector product.}
    \State \texttt{RETURN} $\mathbf{g}_x + \mathbf{g}_{xh}.$
\end{algorithmic}
\end{algorithm}

\section{Experiments}
We conducted a minimum set of experiments to verify that our method works, and is able to solve classification tasks on the level of Neural CDE.

\subsection{Architecture details}

We took the Neural CDE vector field architecture from the examples in the torchcde repository\footnote{\url{https://github.com/patrick-kidger/torchcde/blob/master/example/time_series_classification.py}}. 
It consists of two linear layers: the first one is ReLU-activated, the second is Tanh-activated:
$$
    f_\theta(\mathbf{X}_t, \mathbf{h}_t) = 
    \text{Tanh}(W_2 \text{ReLU} (W_1 \mathbf{h}_t + \mathbf{b}_1) + \mathbf{b}_2).
$$

For the architecture of our method is consistent with that of the Matrix-based Neural CDE. We used a ReLU-activated Elman RNN as the first layer, and a Tanh-activated linear layer as the output:
$$
    f_\theta(\mathbf{X}_t, \mathbf{h}_t) = 
    \text{Tanh}(W_2 \text{ReLU} (W_h \mathbf{h}_t + W_x \mathbf{X}_t + \mathbf{b}_1) + \mathbf{b}_2).
$$
The Jacobians are calculated by hand:
\begin{align*}
    J_x &= \text{Tanh}' \cdot W_2 \cdot \text{ReLU}' \cdot W_x, \\
    J_h &= \text{Tanh}' \cdot W_2 \cdot \text{ReLU}' \cdot W_h,
\end{align*}
where we denote the partial Jacobians of the Tanh and the ReLU activated parts by Tanh' and ReLU'.

During optimization, the autograd engine will need to differentiate $J_x,J_h$, but the second derivative of ReLU is zero.
To mitigate this, we adopt the surrogate gradients trick, introduced for Spiking Neural Networks~\cite{8891809}.
Specifically, we replace the ReLU derivative with the sigmoid function.

\subsection{Numerical Results}
We compared the methods on the CharacterTrajectories dataset~\footnote{\url{https://timeseriesclassification.com/description.php?Dataset=CharacterTrajectories}}.

\begin{table}[tbh]
    \centering
    \caption{Comparison of our Jacobian NCDE to the original matrix-based one. We run the methods on four seeds, and then average over them, also reporting the standard deviation.}
    \label{tab:chartraj}
    \begin{tabular}{ccc}
        \toprule
        Name & Matrix Neural CDE & Jacobian Neural CDE \\
        \midrule
        Metric & 0.93 +- 0.02 & 0.94 +- 0.01  \\
        \# Params & 66K & 33K \\
        \bottomrule
    \end{tabular}
\end{table}

The results are presented in Table~\ref{tab:chartraj}.
As you can see, our Jacobian-based vector field performs as good as the original model, while using half the number of parameters. 

\printbibliography

@article{kidger2020neural,
  title={Neural controlled differential equations for irregular time series},
  author={Kidger, Patrick and Morrill, James and Foster, James and Lyons, Terry},
  journal={Advances in neural information processing systems},
  volume={33},
  pages={6696--6707},
  year={2020}
}

@article{chen2018neural,
  title={Neural ordinary differential equations},
  author={Chen, Ricky TQ and Rubanova, Yulia and Bettencourt, Jesse and Duvenaud, David K},
  journal={Advances in neural information processing systems},
  volume={31},
  year={2018}
}

@article{de2019gru,
  title={GRU-ODE-Bayes: Continuous modeling of sporadically-observed time series},
  author={De Brouwer, Edward and Simm, Jaak and Arany, Adam and Moreau, Yves},
  journal={Advances in neural information processing systems},
  volume={32},
  year={2019}
}

@ARTICLE{8891809,
  author={Neftci, Emre O. and Mostafa, Hesham and Zenke, Friedemann},
  journal={IEEE Signal Processing Magazine}, 
  title={Surrogate Gradient Learning in Spiking Neural Networks: Bringing the Power of Gradient-Based Optimization to Spiking Neural Networks}, 
  year={2019},
  volume={36},
  number={6},
  pages={51-63},
  doi={10.1109/MSP.2019.2931595}}

\end{document}